# Detection of Underwater Multi-Targets Based on Self-Supervised Learning and Deformable Path Aggregation Feature Pyramid Network

Chang Liu

School of Computer Science and Engineering, University of Electronic Science and Technology of China, Chengdu 611731, China

**Abstract.** To overcome the constraints of the underwater environment and improve the accuracy and robustness of underwater target detection models, this paper develops a specialized dataset for underwater target detection and proposes an efficient algorithm for underwater multi-target detection. A self-supervised learning based on the SimSiam structure is employed for the pre-training of underwater target detection network. To address the problems of low detection accuracy caused by low contrast, mutual occlusion and dense distribution of underwater targets in underwater object detection, a detection model suitable for underwater target detection is proposed by introducing deformable convolution and dilated convolution. The proposed detection model can obtain more effective information by increasing the receptive field. In addition, the regression loss function EIoU is introduced, which improves model performance by separately calculating the width and height losses of the predicted box. Experiment results show that the accuracy of the underwater target detection has been improved by the proposed detector.

## 1 Introduction

Underwater target detection plays an important role in ocean activities by human beings, for scientific, commercial or military purposes. With the development of underwater imaging systems and target detection algorithms, underwater target detection has been paid increasing attention. To fulfill the mission of underwater target detection, an underwater vehicle equipped with acoustic or optical sensors would be preferred. However, due to the complexity and uncertainty of underwater environments for instance the light absorption and scattering, usually the obtained underwater images are of low quality, which makes it difficult to accurately detect underwater targets. In addition, in the detection of underwater creatures, the dimensional difference and movement of targets pose another challenge to accurate detection.

Generally, the development of underwater target detection techniques experiences the evolution from traditional target detection based on handcrafted features to ad-vanced target detection based on deep learning. Due to the complexity and uncertainty of underwater environments, traditional underwater target detection faces challenges in achieving robust and accurate detection of underwater targets. Moreover, to improve the detection performance, traditional underwater target detection usually requires tedious feature engineering and extensive prior knowledge of users, which affects the applicability of traditional underwater target detection techniques. During the last decade, deep learning based underwater target detection receives an increasing attention as this approach presents a powerful learning and prediction ability. Li et al. [1] applied fast regions with convolutional neural networks (Fast R-CNN) to the detection and recognition of fish species from underwater images. Sun et al. [2] proposed a transferring deep knowledge CNN for the object recognition in low-quality underwater videos. Zhou et al. [3] applied Faster R-CNN network to the marine organism detection and recognition, combined with data augmentation. Park and Kang [4] used You Only Look Once (YOLO) network to the underwater object recognition, in which the human heuristic approach is applied to increase object classification accuracy. Chen et al. [5] proposed an improved YOLOv4 for the underwater target recognition. Arain et al. [6] proposed a deep convolutional neural network (DCNN) based combining feature-based stereo matching with learning-based segmentation to produce a more robust map in underwater obstacle detection. Jia and Liu [7] proposed a marine animal detection algorithm that combines YOLOv3 and multiscale Retinex with color restore (MSRCR) to improve the accuracy and speed of detection. Qiang et al. [8] proposed a single shot multibox detector (SSD) based underwater target recognition algorithm to improve the accuracy and speed of detection. Zang et al. [9] applied region-based fully convolutional network (R-FCN) to the detection of marine organisms. Liu et al. [10] proposed a fast underwater target recognition by using generative adversarial network (GAN). Li et al. [11] presented a real-time underwater target detection by using an improved YOLO network and transfer learning. Li et al. [12] proposed an underwater target detection algorithm based on an improved YOLOv5 in which path aggregation network (PANet) is combined. Jalal et al. [13] proposed an improved deep neural network for the fish detection and species classification by using temporal information with probabilistic modelling and optical flow. Al Muksit et al. [14] proposed a deep learning-based fish detection model to detect fish in realistic underwater environment. Chen et al. [15] proposed a detection and recognition method of shallow underwater biological target based on YOLOv3 network. Ye et al. [16] presented an improved YOLOv3 based underwater target detection algorithm to solve the problem of faulty and omitted detection. The recent variants of YOLO including YOLOv8 [17], YOLOv9 [18], YOLOv10 [19], and YOLOv11 [20] have also found their application in the underwater detection. Although many YOLO variants are available, it is noted that YOLOv5 and YOLOv8 are usually preferred in practical applications owing to the stability.

Although deep learning based underwater detection has achieved many progresses, some issues need to be further addressed. Firstly, it is difficult to obtain underwater images and the cost of making labeled datasets is high, which limits the practical utility of underwater object detection algorithms. Secondly, due to the particularity of underwater environments, the detection accuracy is affected some effects such as low contrast, mutual occlusion and dense distribution of underwater targets. Thirdly, small underwater targets are difficult to detect in underwater multi-target detection due to the fuzzy underwater environment and the excessive fusion degree of target and background. In this paper, a deep learning based underwater multi-target detection model is proposed. The main contributions include

(1) A self-supervised learning is proposed to help the object detection model learn more general underwater information. In the self-supervised scheme, the SimSiam structure is employed to decrease the effect of negative pairs.

(2) By introducing deformable convolution and dilated convolution, an improved detection model is constructed, in which a regression loss function EIoU is introduced. The improved model can obtain more effective information since the receptive field is increased.

## 2    Self-supervised Learning

To address the issue caused by insufficient annotated dataset, self-supervised learning provides an effective way. Without the need for annotated dataset, self-supervised learning trains models by learning the underlying feature representations of different data adaptively. A self-supervised learning strategy utilizes a pretext task that only requires unlabeled data to construct feature representations with semantic meaning, which can subsequently be used to tackle downstream tasks. Representative self-supervised learning approaches include SimCLR, MoCo, BYOL, and SimSiam. In the study, SimSiam is employed. Its structure is shown as Fig. 1.

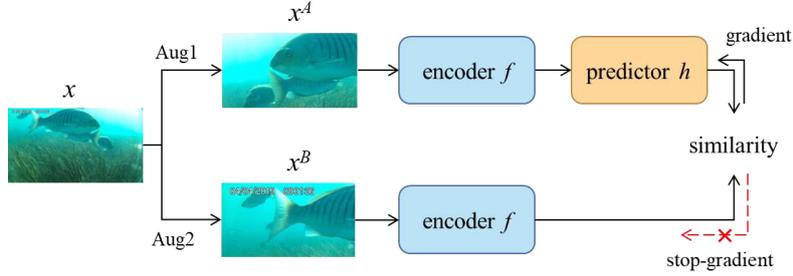

**Fig. 1.** Architecture of SimSiam.

Two images, $x^A$ and $x^B$, can be obtained after random augmentation is performed for a given input image x. An encoder $f(\cdot)$ is then used to process the inputs $x^A$ and $x^B$, yielding $Z^A = f(x^A)$ and $Z^B = f(x^B)$. Then a MLP predictor $h(\cdot)$ is added to one branch, getting $P^A = h(Z^A)$. By feeding the input $x^B$ to the branch that contains the predictor, one can also obtain $P^B = h(Z^B)$. Afterwards, a symmetric loss function is obtained by calculating the negative values of the similarity between predictor results $\{P^A, P^B\}$ and encoder results $\{Z^B, Z^A\}$

$$L_S = \frac{1}{2}D\left(P^A, stopgrad\left(Z^B\right)\right) + \frac{1}{2}D\left(P^B, stopgrad\left(Z^A\right)\right) \tag{1}$$

where $D\left(\cdot\right)$ denotes the negative cosine similarity function; *stopgrad* $(\cdot)$ represents the stop-gradient operation, which means that $Z^A$ or $Z^B$ remains a constant under such an operation.

## 3    Deformable Path Aggregation Feature Pyramid Network

### 3.1    Trident Dilated Convolution

The underwater creatures are characterized by diversity in size, appearance, and clustering, which makes it too difficult to accurately detect. Inspired by the TridentNet and dilated convolution, a trident dilated convolution module (TDConv) is designed to improve the detection accuracy and efficiency. The dilated convolution can improve the receptive field while keeping a small kernel size. As shown in Fig. 2, based on the convolution kernel size $3 \times 3$, the receptive field increases to $7\times7$ by setting a dilation rate $r = 2$, which helps detect underwater targets in complex underwater environments. Considering the difference of size, three dilation rates are adopted in the study. Moreover, to reduce the computation burden due to multiple convolutions, the parameters in three branches are shared by each other, as conducted in TridentNet.

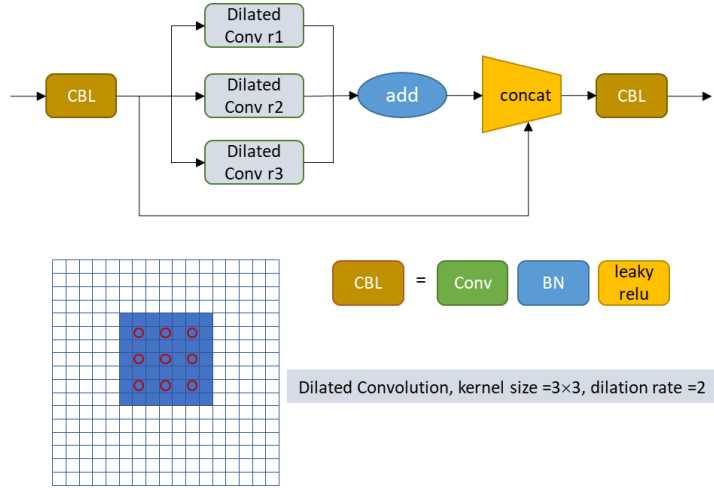

**Fig. 2.** Trident dilated convolution.

Based on the TDConv, a trident dilated spatial pyramid pooling-fast module (TDSPPF) is further designed to improve the detection efficiency and prevent the loss of feature information when multiple pooling is conducted. As shown in Fig. 3, in the SPPF structure, three max pooling layers are connected in series. By combining the TDConv and SPPF, the detection ability for underwater multiscale targets is improved, so is the detection efficiency.

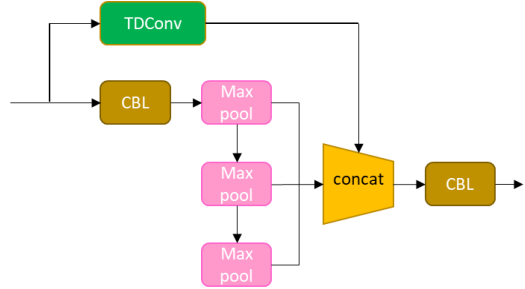

**Fig. 3.** Trident dilated spatial pyramid pooling-fast.

### 3.2    Deformable Path Aggregation Feature Pyramid Network

In YOLOv5, path aggregation feature pyramid network (PAFPN) is adopted in the Neck module to shorten the information path between lower layers and topmost feature. Due to the particularity of underwater environments such as light attenuation and forward/backward scattering, the obtained optical underwater images degrade, represented as low contrast, blurred details, and color distortion, which makes the feature subtraction difficult. In the study, deformable convolution is added to PAFPN to enhance the transformation modeling capability. As pointing out [21], the receptive field and the sampling locations are adaptively adjustable with the target shape and scale by using deformable convolution.

The framework of the deformable path aggregation feature pyramid network (DPAFPN) is shown as Fig. 4. It guarantees the fusion of detail feature and semantic feature. To prevent the loss of information due to the spatial pyramid pooling (SPP) and increase the receptive field, a trident dilated convolution module (TDConv) is introduced to deal with two feature layers. The output feature layers are P3 and P4. To prevent the loss of information due to continuous pooling, a trident dilated spatial pyramid pooling-fast module (TDSPPF) is introduced to deal with the third feature layer. The output feature layer is P5. The outputs from the DPAFPN neck are passed to the heads to detect underwater targets.

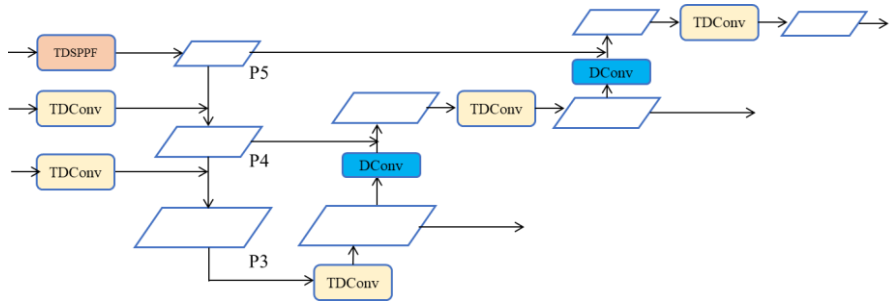

**Fig. 4.** Deformable path aggregation feature pyramid network.

The architecture of the underwater target detection model proposed in the study is shown as Fig.5. The network consists of three modules, i.e. backbone, neck and head. Compared with baseline YOLOv5, the backbone and head components are kept unchanged while the neck is replaced by the proposed DPAFPN. Moreover, the weights in backbone are taken from the pre-training results by using self-supervised learning.

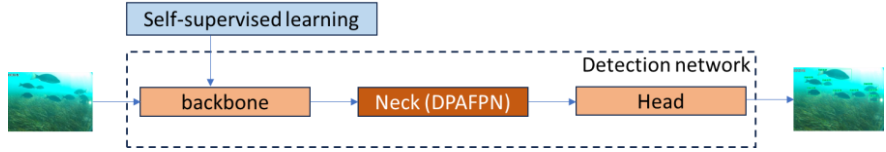

**Fig. 5.** Underwater target detection network.

### 3.3   Loss Function

In the YOLOv5 training, the loss function consists of three parts, i.e. the regression loss, confidence loss, and classification loss. Among the three loss functions, the regression loss is vital to the accuracy of target detection. In the study, a more efficient version of IoU loss, i.e. EIoU loss, is adopted. The EIOU index is defined as

$$EIoU = IoU - \frac{\rho^2(b, b^{gt})}{c^2} - \frac{\rho^2(w, w^{gt})}{c_w^2} - \frac{\rho^2(h, h^{gt})}{c_h^2} \quad (2)$$

where $IoU$ denotes the intersection over union; $b$ and $b^{gt}$ denote the central points of predicted box and target box, respectively; $w$ and $w^{gt}$ denote the widths of predicted box and target box, respectively; $h$ and $h^{gt}$ denote the heights of predicted box and target box, respectively; $\rho(\cdot,\cdot)$ indicates the Euclidean distance; $c$ is the diagonal length of the smallest enclosing box covering predicted box and target box. $c_w$ and $c_h$ denote the width and height of the smallest enclosing box, respectively.

The loss function with respect to (2) is

$$L_{EIoU} = 1 - EIoU \quad (3)$$

Based on the improved $IoU$ loss as (3), the overall regression loss function in an image can be obtained as

$$L_{re} = \sum_{i=0}^{S}(1 - EIoU_i) \quad (4)$$

where $S$ is the number of positive samples in all predicted boxes.

Combining the confidence loss and classification loss, the over loss function in the improved network is determined by

$$L = L_{re} + L_{co} + L_{cl} \quad (5)$$

where the confidence loss function $L_{co}$ is calculated by

$$L_{co} = -\sum_{i=0}^{M}\left[\hat{C}_i \log(C_i) + (1-\hat{C}_i)\log(1-C_i)\right] \quad (6)$$

and the classification loss function $L_{cl}$ is calculated by

$$L_{cl} = -\sum_{i=0}^{S}\sum_{c\in classes}\left[\hat{p}_i(c)\log(p_i(c)) + (1-\hat{p}_i(c))\log(1-p_i(c))\right] \quad (7)$$

where $M$ in (6) is the number of predicted boxes; $C_i$ denotes the prediction confidence or target probability; $\hat{C}_i$ denotes the true category, generally taking 0 or 1; $\hat{p}_i(c)$ is the true probability of class $c$ while $p_i(c)$ is the predicted probability of class $c$.

## 4    Experiments

### 4.1    Dataset

The dataset used for verification is derived from the dataset for the Target Recognition Group of the China 2020 National Underwater Robot Professional Contest (URPC 2020), which includes 4757 images involving four categories of targets, i.e. scallop, echinus, starfish, and holothurian. The total number of the targets is 36100, in which scallop amounts to 6617, echinus 18490, starfish 5794, and holothurian 5199. In the study, to enrich the diversity of target categories, an augmented dataset was constructed based on URPC, which contains 8814 images of nine target categories, i.e. scallop, echinus, starfish, holothurian, fish, turtle, octopus, sea snake, and lobster. Such a dataset extensively covers common organisms in underwater environments. By utilizing this dataset, the robustness and adaptability of proposed detection method can be checked. Moreover, the dataset is split into training and validation sets, with the ratio 9:1. The size of input images are all set as [640, 640].

### 4.2    Underwater Target Detection

The experiment was conducted in a Python, OpenCV, and PyTorch environment, with the respective versions being Python 3.7, OpenCV3.5.2 and PyTorch 1.9.0. The computer operating system was Windows 10, with the following hardware configuration: Processor: Intel Core i5-10600KF (6 cores, 12 threads); Graphics card: NVIDIA GeForce RTX 3080 (10GB). The optimizer in the proposed self-supervised learning adopts Adaptive Moment Estimation (Adam). The encoder consists of the module CSPDarkNet and the module prediction MLP. The projection MLP is also from Simsiam. The downstream target detection network adopts the YOLOv5 detection network. The image augmentation methods refer to horizontal flipping, random cropping, resizing, color distortion (involving brightness, saturation, contrast, and hue), and grayscale conversion. The mask ratio is set as 0.75; the learning rate of the network is set to 0.0001; the batch size is set to 4; the number of epochs for downstream task (object detection) training is set to 100. Two learning strategies are compared, i.e. the supervised learning and the baseline SimSiam. In the training of detector network, the optimizer adopts stochastic gradient descent (SGD) algorithm. The number of epochs is set to 100. To verify the advantages of the proposed detector in the detection of underwater targets, several other detectors are compared, including YOLO-family algorithms (YOLOv5, YOLOv7), Swin Transformer, and a representative two-stage detector (Faster RCNN). The comparison results are shown as Fig. 6. It is noted that in the detection results the detected targets are labelled with a box containing category and probability while the missed targets are highlighted with red boxes.

Quantitative comparison among different detection models is given in Table 1. In Table 1, the mean average precision (mAP) for all kinds of underwater creature and average precision (AP) for each category are calculated, under the $IoU = 0.5$ case. The best precisions are denoted by bold font.

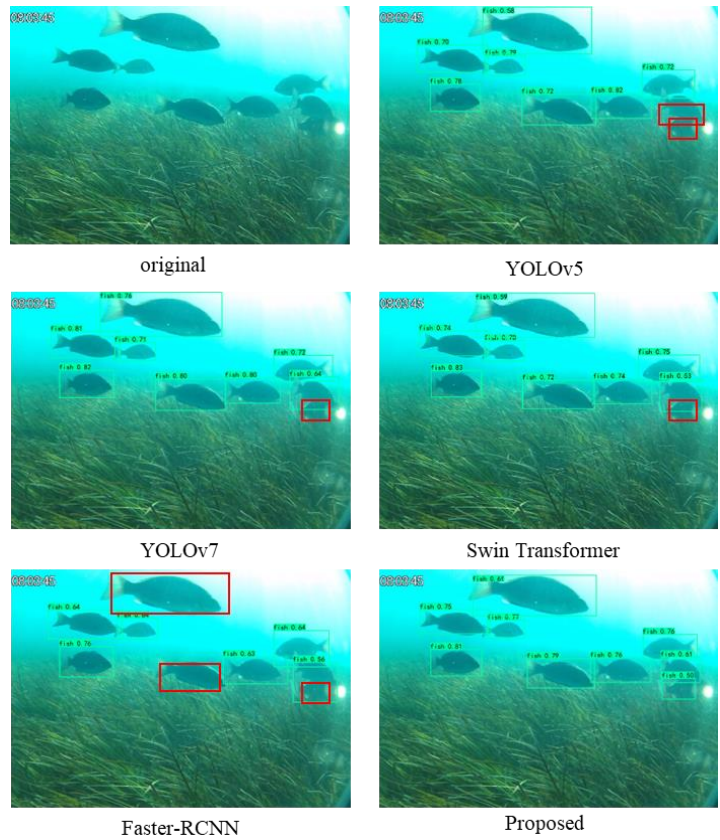

**Fig. 6.** Detection results.

Table 1. Comparison of different detection models in terms of mAP and AP.

| Model | mAP(%) | Starfish | Scallop | Holothurian | Fish | Lobster | Octopus | Sea snake | Turtle | Echinus |
|---|---|---|---|---|---|---|---|---|---|---|
| YOLOv5 | 71.41 | 82.46 | 53.16 | 35.34 | 88.85 | 87.03 | 29.15 | 82.65 | 94.16 | 89.91 |
| YOLOv7 | 73.44 | 82.36 | 49.76 | 30.98 | 87.65 | **92.61** | 43.94 | 90.64 | 92.20 | **90.78** |
| S-T | 72.81 | 80.07 | 45.06 | 32.37 | 88.37 | 91.50 | 45.58 | 90.34 | 94.09 | 87.92 |
| F-R | 42.75 | 2.61 | 0.44 | 2.07 | 56.12 | 79.64 | 48.51 | 70.69 | 68.94 | 55.77 |
| Proposed | **77.42** | **83.49** | **55.64** | **39.30** | **89.75** | 90.83 | **60.54** | **92.28** | **94.74** | 90.23 |

As can be seen from the quantitative comparison results, the proposed detection model achieves the best mAP over the other detection models. In terms of AP for each category, the proposed model also performs excellently. Especially for the octopus that usually deforms due to the underwater environments, the improvement of precision by using the proposed detection model is significant, compared with baseline YOLOv5 and the other detection models. Except for the cases of lobster and echinus, the proposed model gains the best APs for the other categories. Nevertheless, the detection of lobster and echinus performs well, under the proposed model.

Table 2 lists the comparison between supervised learning and the baseline SimSiam. The detection network is taken as YOLOv5. From the comparison results, it can be seen that self-supervised learning improves the detection accuracy.

Table 2. Comparison of supervised learning and self-supervised learning.

| Model | supervised learning | SimSiam learning |
|---|---|---|
| mAP(%) | 71.41 | 72.06 |

### 4.3 Ablation Experiments

In the proposed detection model, the main contributions refer to three modules, i.e. trident dilated convolution module (TDConv), trident dilated spatial pyramid pooling-fast module (TDSPPF), and deformable path aggregation feature pyramid network (DPAFPN). In structure, the detection model with DPAFPN is more complicated than the model with TDSPPF, as can be recognized from Fig. 4. To verify the effectiveness of each module, ablation experiments are conducted and compared with the baseline YOLOv5. The mAP and mAR comparison under different IoU values is shown in Fig. 6, in which model a denotes the model consists baseline YOLOv5 and TDSPPF; model b denotes the model consists of baseline YOLOv5 and DPAFPN; model c consists of YOLOv5, TDSPPF, and DPAFPN; while the proposed model consists of the baseline YOLOv5 and all modules, i.e. TDSPPF, DPAFPN, and TDConv.

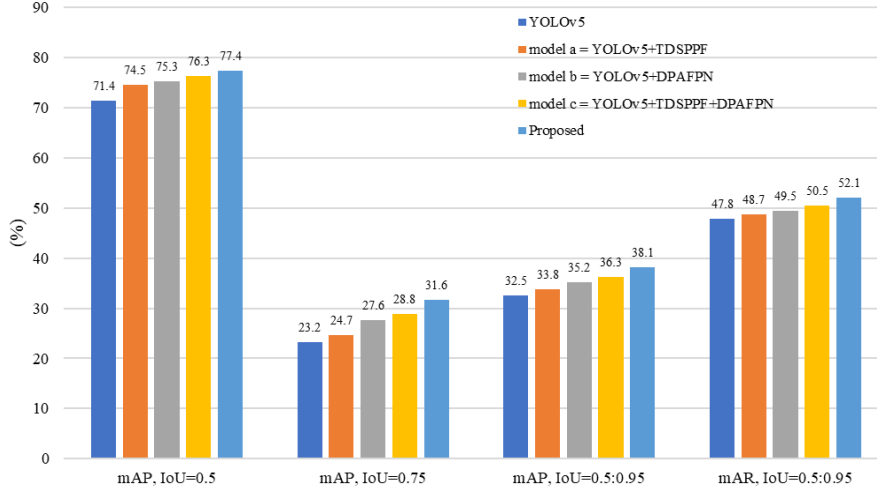

As can be seen from the comparison results in Fig. 6, the designed modules implanted into the YOLOv5 do improve the detection accuracy. The most structurally complicated model, i.e. the model consisting of baseline YOLOv5, TDConv, TDSPPF, and DPAFPN, achieves the best performance comparing with the other stripped models.

## 5　　Conclusions

A deep learning model is proposed to the underwater target detection. Considering the characteristics of underwater creatures and environments, a self-supervised learning strategy and improved YOLOv5 are proposed. The proposed self-supervised learning aims to address the difficulty in obtaining underwater datasets and the difficulty with annotation of underwater images. The proposed YOLOv5 model aims to improve the detection accuracy and reliability when the underwater environment results in degradation of underwater images and the underwater creatures are characterized by diversity in size, appearance, and clustering, which makes it too difficult to accurately detect. To these ends, some measures are taken and verified, including

(1) Self-supervised learning is employed, which helps the target detection model capture more effective feature representation under complicated underwater environments.

(2) In the proposed detection model, three modules are designed, including trident dilated convolution module (TDConv), trident dilated spatial pyramid pooling-fast module (TDSPPF), and deformable path aggregation feature pyramid network (DPAFPN). These modules help the target detection model subtract more effective topmost feature by improving the receptive field. Moreover, to improve the detection accuracy, an efficient IoU loss function is proposed.

It is noted that in the study the degradation of underwater images is not dealt with, which affects the detection accuracy. This is can be seen from the mAP results in the tables. Undoubtedly, enhancement of underwater images does improve the detection accuracy. In future work, the underwater image enhancement will be taken into account before target detection is conducted. Moreover, the introduction of multiple modules might increase computational complexity. Therefore, in the next work efforts will be devoted to both mAP and the model complexity to evaluate the practicality and suitability of the detection model.

**Acknowledgments.** The authors appreciate the anonymous review for their help in improving the quality of the paper.

**Disclosure of Interests.** The authors have no competing interests to declare that are relevant to the content of this article.